\documentclass{article}
\usepackage[english]{babel}
\usepackage{times}
\usepackage[letterpaper,top=2cm,bottom=2cm,left=3cm,right=3cm,marginparwidth=1.75cm]{geometry}
\usepackage{soul}
\usepackage{xcolor} 
\usepackage{amsmath}
\usepackage{amsfonts} 
\usepackage[notransparent]{svg} 
\usepackage{graphicx}
\usepackage{subcaption}
\usepackage[colorlinks=true, allcolors=blue]{hyperref}
\usepackage{algorithm}
\usepackage{algpseudocode}
\usepackage{booktabs} 
\usepackage{multirow}
\usepackage{makecell}
\usepackage{array} 
\usepackage{mathtools}
\usepackage[normalem]{ulem}

\title{TruKAN: Towards More Efficient Kolmogorov-Arnold Networks Using Truncated Power Functions}
\author{Ali Bayeh\thanks{ORCID: \href{https://orcid.org/0009-0006-6402-9402}{0009-0006-6402-9402}} \quad\quad
Samira Sadaoui\thanks{ORCID: \href{https://orcid.org/0000-0002-9887-1570}{0000-0002-9887-1570}} \quad\quad
Malek Mouhoub\thanks{ORCID: \href{https://orcid.org/0000-0001-7381-1064}{0000-0001-7381-1064}} \\
Department of Computer Science, University of Regina, SK, Canada\\
\texttt{\{alibayeh, sadaouis, mouhoubm\}@uregina.ca}

\definecolor{brightmaroon}{rgb}{0.76, 0.13, 0.28}
}
\date{} 

\begin{document}
\maketitle
\begin{abstract}
To address the trade-off between computational efficiency and adherence to Kolmogorov-Arnold Network (KAN) principles, we propose TruKAN, a new architecture based on the KAN structure and learnable activation functions. TruKAN replaces the B-spline basis in KAN with a family of truncated power functions derived from k-order spline theory. This change maintains the KAN's expressiveness while enhancing accuracy and training time. Each TruKAN layer combines a truncated power term with a polynomial term and employs either shared or individual knots. TruKAN exhibits greater interpretability than other KAN variants due to its simplified basis functions and knot configurations. By prioritizing interpretable basis functions, TruKAN aims to balance approximation efficacy with transparency. We develop the TruKAN model and integrate it into an advanced EfficientNet-V2–based framework, which is then evaluated on computer vision benchmark datasets.  To ensure a fair comparison, we develop various models: MLP-, KAN-, SineKAN  and TruKAN-based EfficientNet frameworks and assess their training time and accuracy across small and deep architectures. The training phase uses hybrid optimization to improve convergence stability. Additionally, we investigate layer normalization techniques for all the models and assess the impact of shared versus individual knots in TruKAN. Overall, TruKAN outperforms other KAN models in terms of accuracy, computational efficiency and memory usage on the complex vision task, demonstrating advantages beyond the limited settings explored in prior KAN studies. 
\end{abstract}


\section{Introduction}
The renewed interest in Kolmogorov–Arnold Networks (KANs) has highlighted their substantial potential, prompting investigations in diverse domains, such as computer vision \cite{visionKAN}, finances \cite{FinanceKAN, FinanceKAN2}, biomedical applications \cite{biomedicalKAN}, biological tasks \cite{GenomicKAN} and complex reinforcement learning environments \cite{OurPpoKAN, OurSacKAN}.  In KANs, evaluating the learnable activation functions (often represented as splines) is computationally expensive because it relies on the recursive de Boor–Cox algorithm \cite{deboor1978}.  As noted in \cite{KAN}, KANs can typically train up to 10 times slower than MLPs with equivalent parameter counts. Training becomes even more difficult for high-dimensional problems and deep architectures: memory and computation grow quickly, gradients become noisier, and optimization becomes unstable. Improving the speed of KANs while preserving their expressive power, accuracy, and training stability at scale is essential to make them practical for large applications. Although several studies have proposed improvements to KANs, most evaluations have been limited to synthetic data and simple benchmarks such as MNIST.  
\medskip 

Inspired by recent advancements in KANs \cite{KAN,MultKAN} and based on the spline theory, our study introduces a new variant of KAN, called TruKAN, to address the computational inefficiencies and scalability issues of KANs. TruKAN architecture preserves the canonical KAN topology and learnable activations but replaces the B-spline algorithm with a family of truncated power functions. This substitution maintains the expressiveness of KANs while improving performance and interpretability. Each TruKAN layer combines a truncated power component and a polynomial component. We also propose two variants of TruKAN that
differ in how the knots are used: shared knots over outputs or individual knots per output. MLPs rely heavily on nonlinear activation functions (by the universal approximation theorem), which enables large representational capacity but compromise interpretability. The integration of truncated power bases with polynomial terms in TruKAN addresses this limitation, yielding to a more interpretable framework compared to KANs and variants, such as SineKAN \cite{SineKAN} and FourierKAN \cite{FourierKAN}. The latter typically incorporate nonlinearities (like SiLU in KAN) or exhibit intrinsic nonlinear behavior, which can make them less interpretable. By prioritizing interpretable basis functions, TruKAN aims to balance approximation efficacy with transparency.

From a theoretical perspective, truncated power bases are a sub-optimal choice for function approximation compared to B-splines \cite{deboor1978}, because they have difficulty capturing functions with abrupt discontinuities or sharp transitions. However, in practical settings, the high degree of locality afforded by B-splines in KANs may not be essential, and truncated power bases can perform adequately within the hierarchical structure of KANs, potentially simplifying implementation without significant loss in performance. However, the exponential growth or decay characteristics of truncated power functions require specialized training strategies that constrain the values of the coefficients and prevent gradient explosion. This research focuses on developing and evaluating these techniques, exploring the empirical viability of TruKAN, and identifying its advantages over the counterparts. Our study includes a theoretical analysis of basis function properties, empirical validation through benchmarks and methodological advancements in training stabilization. 
\medskip 

Our paper evaluates TruKAN model using computer vision datasets under different architectural configurations. Vision tasks are important because they support numerous critical real-world applications, such as autonomous vehicles, medical imaging and environmental monitoring. We first design an advanced deep architecture based on EfficientNet-V2 \cite{Efficientnetv2}. In this architecture, we employ different classifiers: MLP-based (with or without layer normalization), KAN-based (with or without layer normalization), SineKAN-based (with or without layer normalization), and TruKAN-based (with or without layer normalization, and with individual or shared knots). We also utilize a comprehensive data augmentation pipeline for each dataset to improve the training process. For each dataset, we develop and assess two MLP models, two KAN models, two SineKAN models and four TruKAN models for each of two configurations:  a small architecture and a deep architecture. Our goal is to compare different network configurations based on training time and other testing quality metrics.
For all experiments, we keep the number of classifier parameters similar across methods to ensure that any performance differences is interpreted as algorithmic rather than due to the network capacity. Overall, TruKAN outperforms other KAN models in accuracy, time and memory usage on the complex vision task, demonstrating advantages beyond the constrained settings explored in prior KAN studies. 

\medskip
Section 2 describes the foundations and underlying architecture of KANs. Section 3 reviews recent studies on different improvements of KAN. Section 4 details the theory, configuration and components of our new TrunKAN method. Sections 5 describes our experimental setup, including the training optimization, the selected computer vision datasets and data augmentation pipeline.  Sections 6 presents our EfficientNet-based model configuration, regularization techniques, and hyper-parameters tuning. Section 7 evaluates and compares the performance of numerous models under different architecture sizes. Finally, Section 8 concludes the paper and outlines directions for future research.
\section{Background on KAN Architectures}
Inspired by the Kolmogorov-Arnold representation theorem, KAN models \cite{KAN,MultKAN} differ from traditional MLPs by replacing fixed activation functions with learnable ones.
Instead of using scalar linear weights, KANs learn univariate edge-wise functions (often parameterized as splines), which makes the parameters interpretable and enables very compact KANs to match or exceed the accuracy of much larger MLPs \cite{KAN,MultKAN}. 
KANs are better suited for visualization, interpretability, and subsequent symbolic extraction, since their learned edge functions have a structured and inspectable form.
These properties make KANs attractive for scientific discovery and interdisciplinary research, bridging the gap between ML and domain science.

Recent advancements in KANs \cite{KAN,MultKAN} utilize the B-spline method as the basis for activation functions, creating smooth, trainable components. The KAN model consists of a stack of KAN layers, each of which follows the Kolmogorov-Arnold theorem. Each KAN layer is composed of two parts: 1) an input component ($\varphi_{0,j,i}$) that decomposes according to a grid size, and 2) an output component ($\varphi_{1,j,i}$) that computes the aggregated final output based on the results of the decomposition, as follows \cite{KAN}:
\begin{equation}
  \begin{split}
    x_{l,j} = \sum_{i=1}^{n_l} \hat{x}_{l-1,j,i} = \sum_{i=1}^{n_l} \varphi_{l-1,j,i} (x_{l-1,i}) \:\quad \quad \quad \quad \quad\\\\
    \varphi_{l,j,i}(x_{l,i}) = \sum_j c_j B_{l,j,i}(x_{l,i})  \:\quad \quad \quad \quad \quad \quad \quad \quad \quad
  \end{split}
  \begin{array}{l}
    l=1,\dots,L \\
    i=1,\dots,n_l \\
    j=1,\dots,n_{l+1}
  \end{array}    
\end{equation}

\noindent here $L$ denotes the maximum number of layers, $n_l$ the number of neurons in each layer $l$, $x_{l,i}$ the activation value of neuron $i^{th}$ in $l^{th}$ layer;  $\varphi_{l,j,i}$ represents the activation function between $i^{th}$ neuron of $l^{th}$ layer and $j^{th}$ neuron of  $\left({l+1}\right)^{th}$ layer.

\begin{figure}[ht]
    \centering
    \includesvg[inkscapelatex=false, width=1\textwidth]{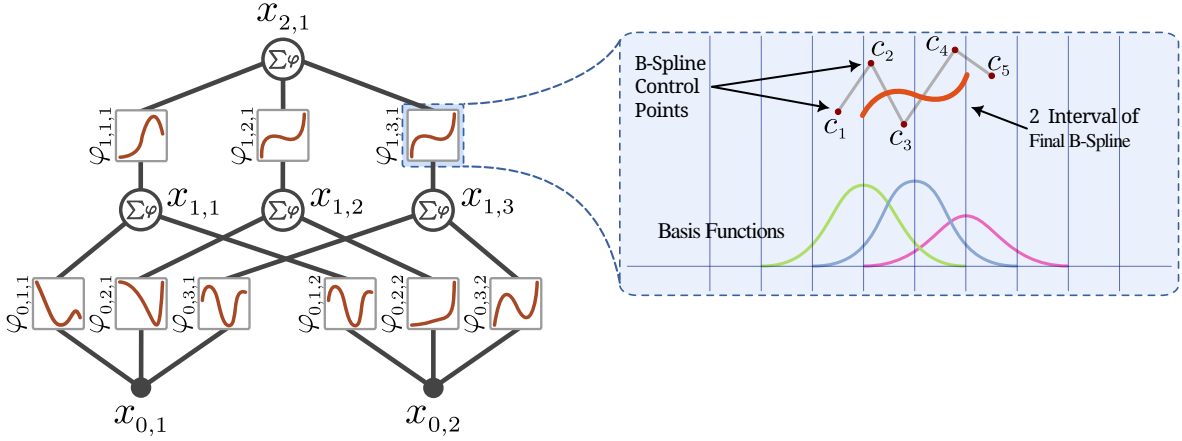}
    \caption{An example of KAN architecture of form $\varphi_\mathbf{t}(x) = \sum_j c_j B_{j,\mathbf{t}}(x)$ according to  \cite{KAN}. The left side presents the notation for activations through the network. The right side depicts an activation function modeled with B-spline. The parameterization of the activation function allows adaptive switching between coarse and fine grid resolutions.}
    \label{fig:KAN_overview}
\end{figure}

Figure \ref{fig:KAN_overview} illustrates an example of the KAN model with two inputs $(x_{0,1}, x_{0,2})$ and a single output $(x_{2,1})$. This architecture implements the theoretical concept by learning many univariate mappings $\varphi$ (one per input $\times$ output “edge”), then summing them to get the final output(s). As depicted on the right side, the grid has fixed intervals and holds the knot locations and coefficients. The knots are the learnable control points, which usually are positioned between knots. A set of coefficients holds the learnable spline parameters that determine the curve shape. In practice, splines alone may be less effective at capturing global data trends. They often require many knots, which increases the number of parameters, and can cause optimization and extrapolation issues. To address these issues, the KAN architecture incorporates a parametric nonlinear basis component (such as a SiLU) to capture global patterns. While the theorem imposes no restrictions on the parametrization and only ensures the existence of univariate maps, adding a global component is consistent with the theorem.
\section{Related Works}
The great potential of KAN architectures reignited attention toward them, prompting numerous studies to address their limitations and enhance their performance. For instance, Fourier analysis provides a natural mathematical foundation for KAN extensions. The Fourier representation of functions as sums of sines and cosines permits arbitrarily close approximation of sufficiently regular periodic functions. Building on this idea, the authors of \cite{FourierKAN} replaced the original B-spline components in KAN architectures with Fourier-based transformations. The resulting model, FourierKAN, is used for feature transformation in Graph Collaborative Filtering (GCF) to improves GCF's representational capacity (by introducing richer, learnable spectral and edgewise transforms). Empirical results on public recommendation benchmarks, such as MOOC and Amazon Games, indicate that FourierKAN reliably trains with simple regularization (message/node dropout) and consistently outperforms baseline MLP models. However, there are some open questions with the proposed approach. For example, the LightGCN study \cite{LightGCN} has previously argued that feature transforms can degrade CF training. Therefore, the benefit of reintroducing a learnable transform depends on careful design and may be dataset dependent. So, while FourierKAN-GCF’s reported gains come from a small set of benchmarks, a broader validation is needed to assess its general applicability.

Another study \cite{WaveKAN} proposed Wav-KAN, which includes wavelet functions into KAN architecture and replace the usual per-edge spline parameterizations with wavelet-based univariate functions. Incorporating each edge with multiresolution basis functions (continuous and discrete wavelet transforms) enable Wav-KAN to capture both low- and high-frequency structures in the inputs. Empirical results show that the multi-resolution properties of wavelets enable Wav-KAN to capture complex, localized data patterns more effectively and offer improved interpretability compared to MLP and B-spline–based KANs. Although Wav-KAN preserves the per-edge univariate form, the performance and inductive bias of the model depend sensitively on the chosen base wavelet, scale ranges, and discretization strategy.

Building on a similar principle, SineKAN, introduced in \cite{SineKAN}, simplifies the Fourier representation by omitting cosine components and parameterizing KAN with learnable frequencies. This modification reduces the number of parameters by half, while retaining the sinusoidal basis for the function approximation. Evaluated on the MNIST dataset, SineKAN achieved performance comparable to B-spline-based KANs while significantly improving training time. However, using the sine function as a replacement for a spline component can be problematic. While splines can capture local details, sine functions are less localized, which can make interpretation of edge functions more difficult in practice. While the SineKAN architecture is faithful to the theory, changing the internal form from localized spline plots to global spectral coefficients affects the model interpretability and requires further analysis.

The BSRBF-KAN paper \cite{BSRBFKAN} constructed a hybrid KAN architecture that augments the B-Spline components with radial basis functions (RBFs). This combination aims to increase representational flexibility and improve stability during training. Through multiple trials on MNIST and Fashion-MNIST datasets, the study concluded that BSRBF-KAN achieved faster and more reliable convergence than several competing KAN variants, including EfficientKAN, FastKAN, FasterKAN and GottliebKAN. These results suggest that integrating B-splines with other mathematical function approximation principles can positively impact optimization dynamics and robustness. However, mixing bases increases the model’s hyperparameter burden and introduces redundancy between components (e.g. radial and spline) that requires regularization or pruning to prevent overfitting. Additionally, the global support of other components (such as RBFs) can affect the localized interpretability of spline-KANs and requires constraints to mitigate this issue.

The study in \cite{LeanKAN} targets the practical shortcomings of a previous study, MultKAN \cite{MultKAN}. LeanKAN is introduced as a sparser model than MultKAN, where the footprint of hyperparameters is reduced while preserving the multiplicative capabilities when needed. The results show that, empirically, LeanKAN serves as a one-to-one drop-in substitution for AddKAN and MultKAN layers. Despite its compact structure, LeanKAN often matches or substantially outperforms larger MultKAN instances on canonical KAN toy problems and KAN-ODE dynamical systems. However, the current evaluation focuses on toy benchmarks and dynamical systems. Thus, a broader validation is needed to fully characterize when and why the leaner design is preferable (e.g., on high-dimensional vision, genomics or NLP tasks).

The authors of \cite{MetaKAN} proposed a new model to improve the memory efficiency of KANs. They constructed a compact meta-learner to address the large memory footprint caused by per-edge learnable univariate functions. This meta-learner cooperates with the KAN model to generate KAN weight functions on the fly. Then, the meta-learner and target KAN were trained end-to-end, demonstrating that this hyper-network style substantially lowers the number of directly optimized parameters and overall memory usage,  while achieving equal or better accuracy on benchmarks, such as symbolic regression, PDE solving, and image classification. However, the method introduces an additional inductive bias and architectural bottleneck by forcing per-edge functions to be produced from a shared generator. This can limit the flexibility of a fully parameterized KAN and requires careful design and tuning of the meta-learner to prevent underfitting or instability during meta-optimization.

The recent study \cite{PowerMLP} introduced PowerMLP, an MLP-architecture that replaces KAN’s explicit per-edge spline parameterization with compact, node-wise polynomial bases combined with linear mixing. While it proved expressivity equivalence to KAN families over bounded domains, its design yields substantial computational advantages over KANs. However, shifting from edge-wise spline coefficients to node-wise power bases entails an  interpretability trade-off. Although PowerMLP offers faster training and inference and simpler implementation, a trained PowerMLP does not present its learned mapping in the human-readable spline form found in KANs. To address this issue, \cite{PowerMLP} supplies constructive conversion lemmas showing how PowerMLP can be mapped back to an equivalent KAN representation. However, this conversion is nontrivial (it involves basis changes and dense coefficient expansions) and may increase parameter complexity and numerical conditioning issues. Evaluations on the Knot Theory dataset and Function Fitting Task show that PowerMLP generally performs better than KANs and can be trained much faster, but it fails to atain high performance in complicated tasks, such as image classification and NLP \cite{PowerMLP}.
\section{TruKAN Description}
\subsection{An Overview}
\begin{figure}[h!]
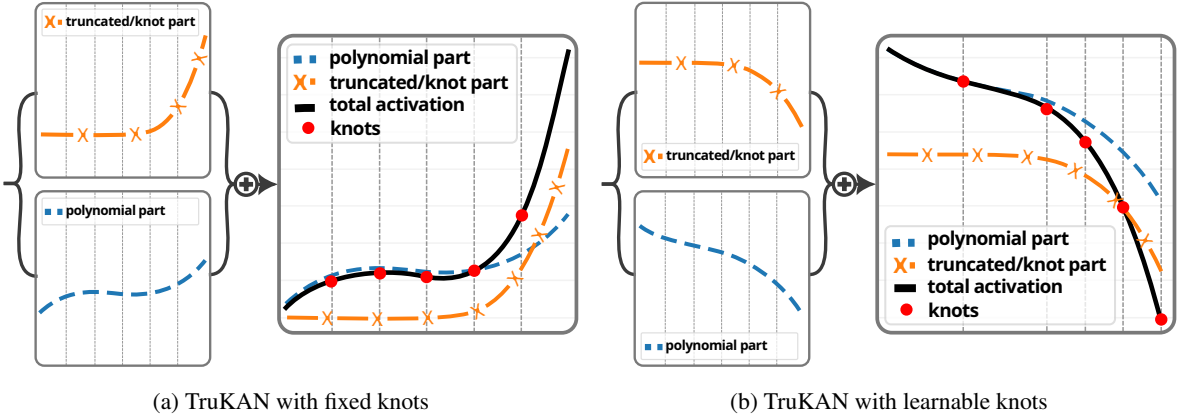

    \centering
    
    \begin{subfigure}[b]{0.49\textwidth}
        \centering
        \includesvg[inkscapelatex=false, width=\textwidth]{images/TruKAN_fixed} 
        \caption{TruKAN with fixed knots}
        \label{fig:TruKAN_fixed}
    \end{subfigure}
    \hfill
    \begin{subfigure}[b]{0.49\textwidth}
        \centering
        \includesvg[inkscapelatex=false, width=\textwidth]{images/TruKAN_learnable}
        \caption{TruKAN with learnable knots}
        \label{fig:TruKAN_learnable}
    \end{subfigure}
    
    \caption{An example of the structure of a TruKAN layer: a truncated power component and a polynomial component: (a) shows a layer with fixed knots, meaning all the knot locations are defined based on equal intervals (here 6) and remain fixed during training; (b) shows a TruKAN layer with learnable knots, meaning the position of the knots will change during training (here 6 variable intervals); all the knot locations are adjusted during training to guarantee an ordered, positive and incremental positioning.}
    \label{fig:TruKAN_arch}
\end{figure}

Figure \ref{fig:TruKAN_arch} illustrates an overview of the internal composition of a TruKAN layer. Each layer combines a truncated-power component with a polynomial component to capture complementary aspects of the mapping. The left panel depicts the fixed-knot variant in which knot locations are placed at equal intervals and remain constant throughout training. In this configuration, the truncated-power basis supplies localized, piecewise nonlinearity while the polynomial term provides a global, low-order trend, producing a stable and interpretable basis with a small number of learnable coefficients. The right panel shows the learnable-knot variant, where knot positions themselves are treated as trainable parameters and are updated during optimization subject to constraints that preserve positivity, ordering, and incremental spacing. The learnable knots increase the layer’s local adaptability and approximation power by concentrating basis flexibility where the data requires it. 

The key idea of KAN is to replace the fixed activation functions (e.g., ReLU) on the edges of a computational graph with learnable univariate functions, parameterized as splines. Specifically, KAN layers implement functions of the form $\sum_{j} c_j B_{j,k,\mathbf{t}}(x)$, where the coefficients $c_j$ are the learnable parameters. 
The theoretical equivalence between the B-spline basis and truncated power basis (formalized via divided differences), guarantees that the spline space used in KAN is the same as that spanned by the $k$-th order truncated power functions \cite{deboor1978}. 

Similar to the KAN architecture, each TruKAN layer consists of two components: 1) a spline component based on truncated power functions, and 2) a basis component composed of polynomial terms. The layer's output is the sum of these two components.
Although KAN and TruKAN have identical representations, they differ at the theoretical level.
The B-spline approach in KAN uses SiLU terms as auxiliary global variables to enhance B-spline locality. In contrast, TruKAN relies on global polynomial terms for spline approximation. Mathematically, the combination of global polynomial terms and truncated power terms in TruKAN corresponds to the B-spline terms in KAN. Thus, TruKAN lacks an equivalent to the auxiliary SiLU terms in KAN. However, in practice, the polynomial terms in TruKAN can compensate for the SiLU terms' contribution in KAN. The following sections compare the polynomial terms with the SiLU terms and the truncated power terms with the B-spline terms.
\subsection{TruKAN Spline Component}
According to spline theory \cite{deboor1978}, each spline with degree $k$ and with knots $\mathbf{t} =\{t_j\}_{j \in \mathbb{Z}}$ is exactly the linear span of these truncated power functions. In the de-Boor method \cite{deboor1978}, a degree-$k$ B-spline basis $B_{i,k}(x)$ is defined recursively over a non-decreasing knot sequence ${t_j}$. Each $B_{j,k}$ has a compact support in $[t_j, t_{j+k+1}]$ and $C^{k-1}$ is continuous at each knot. Unlike B-splines, which are compactly supported, each truncated power function is not compactly supported and its vanishes for $x < t_j$ but grows polynomially for $x \geq t_j$. Nevertheless, any linear combination of these basis functions expressed as \cite{deboor1978}:

\begin{equation}
\label{equ:trunc_spline}
S_\mathbf{t}(x) = \sum_j c_j (x - t_j)_+^k
\end{equation}

\noindent forms a spline of degree $k$ with knots at $t_j$. Here, $S(x)$ denotes the spline function, and $c_j$ are the coefficients that determine the contribution of each truncated power basis function to the spline $S(x)$. This formulation ensures that the resulting function $S(x)$ lies within the same function space as B-splines of degree $k$ and provides an alternative for spline representation.

The equation \ref{equ:trunc_spline} shows that there is an invertible linear map between coefficients in the B-spline basis ($B_{i,k}$)  and coefficients in the truncated power basis. So, any function that can be represented as $\sum a_i B_{i,k}(x)$ can equivalently be written (with different weights) as $\sum c_i (x-t_i)^k_+$ \cite{deboor1978}. Regarding backpropagation, which relies heavily on differentiation, the derivative of B-spline of order $k$ remains a linear combination of B-splines of order $k-1$. On the other hand, the derivative of a $k$-th order truncated power function yields a truncated function of order $k-1$ multiplied by $k$. In general, for $k\ge 0$ and $m\le k$, the $m$-th derivative defines as \cite{deboor1978}:

\begin{equation}
    \frac{d^m}{dx^m}(x-t_0)_+^k = k(k-1)\cdots(k-m+1)(x-t_0)_+^{k-m}
\end{equation}

So that all derivatives of order $m<k$ remain continuous at $x=t_0$ (indeed they vanish at the knot with the expected algebraic order), while the $k$-th derivative reduces to the step-like function $k!\:1_{(x \ge t_0)}$, exhibiting a jump of magnitude $k!$ at the knot. 

While B-splines and truncated power bases span the same mathematical space, their computational properties are dramatically different. The computational complexity of evaluating B-splines of order $k$, using de Boor's algorithm, is $O(k^2)$ per knot interval \cite{deboor1978}, as higher-order splines are recursively constructed from lower-order ones. This scaling is a direct consequence of their local support property, wherein each basis function is non-zero only over an interval spanned by $k+1$ knots. This locality ensures that any point evaluation requires the summation of at most $k+1$ basis functions, and the recursive algorithm efficiently builds higher-order splines from lower-order components.

In contrast, the computational complexity of truncated power basis is $O(G + k)$ per evaluation point \cite{Schumaker2007}, where $G$ denotes the number of knots. This linear scaling with the number of knots arises from the global support nature of the basis where for a given input $x$, all truncated power functions associated with knots before $x$, will contribute to the final sum. Consequently, as the grid resolution $G$ increases, the computational cost grows proportionally, becoming prohibitively expensive for high-resolution models. 

In theory, the computational complexity of truncated power basis functions presents significant challenges for high-resolution knots. However, in practice (as discussed in the experiment section), such dense grid systems is not required, and the TruKAN model performs well with a small number of knots. Therefore, the computational complexity of truncated powers is approximately $k^2$ similar to KAN. 

Another significant divergence between the B-spline and truncated power function bases is their numerical stability.
The piecewise polynomial structure and local support of B-spline basis functions ensure knot and interval stability. In contrast, the truncated power basis incorporates high-order polynomial terms directly, which can introduce numerical instability and complicate the training process.

To mitigate numerical instability associated with truncated power basis functions, we can impose constraints on the output range. Such constraints can be implemented through bounded non-linear activation functions or value clamping techniques. However, while these approaches will resolve the numerical issues, they will compromise the model's interpretability.

The resulting architecture would resemble a MLP in terms of both interpretability and visualization capabilities. The primary method for preserving interpretability while addressing numerical instability in truncated power functions is normalization. By normalizing values to a bounded range during training, one can maintain numerical stability without sacrificing the transparent function representation that characterizes KANs. The theoretical analysis indicates that instability tends to increase with higher spline orders. Thus, numerical instability may arise in higher-order basis functions, even when normalization is applied. However, in practice, low-order bases with $k \le 3$ typically yield stable training and satisfactory performance.

\subsection{TruKAN Basis Component}
The KAN paper \cite{KAN} proposed the basis function as a fixed, non-linear, element-wise mapping from the input space to the space defined by the function. The authors used standard SiLU as a fixed, nonlinear base function shared across all inputs and outputs. This function provides a smooth, differentiable global mapping that ensures gradient continuity and facilitates stable optimization. The SiLU function behaves approximately linearly for large positive values and smoothly saturates for negative ones, endowing the layer with local curvature and bounded gradients. Its derivative is given below:

\begin{equation}
\text{SiLU}(x) = x\sigma(x) = x (\frac{1}{1 + e^{-x}})  
\end{equation}

\begin{equation}
\frac{\text{SiLU}(x)}{dx} = \sigma(x) + x \: \sigma(x) (1 - \sigma(x))
\end{equation}

\noindent SiLU is continuous and strictly positive for all inputs $x$, which ensures it is monotonic and infinitely differentiable. This smoothness guarantees that layer outputs remain globally continuous and differentiable, even when combined with piecewise spline components. In the broader architecture, this property enhances training stability by avoiding abrupt changes in curvature or gradient discontinuities. However, SiLU introduces a fixed nonlinearity bias, as its curvature and saturation behavior are uniform across all input dimensions and independent of the data distribution. As a result, although the KAN achieves universal approximation capability through the combination of SiLU and B-splines, the predefined shape of SiLU may reduces analytical transparency, as individual contributions of coefficients cannot be easily isolated.  In TruKAN, this fixed nonlinearity is replaced with a polynomial base: 

\begin{equation}
\mathbf{Poly}(x) = \sum_{r=0}^{k} a_r x^r
\end{equation}

\noindent whose the $m$-th derivative is as follows:

\begin{equation}
\frac{\mathbf{Poly}^{(m)}(x)}{dx} = \sum_{r=m}^{k} \frac{r!}{(r - m)!} a_r x^{r - m}
\end{equation}

\noindent and thus maintain continuous differentiability up to order $k$. Unlike SiLU, the polynomial coefficients $a_r$ directly contribute to the shaping of both the global and local curvatures. This design closely aligns with classical spline theory \cite{deboor1978}, in which the polynomial part provides the continuous backbone of the function and the truncated-power terms account for local deviations at knots. The final approximation remains smooth within each polynomial region while allowing transparent, piecewise modification of curvature. As a result, TruKAN practically achieves approximation power similar to the KAN, but through a more decomposable and interpretable hierarchy. However, this interpretability comes at the cost of numerical instability issues (similar to the truncated power bases) and the aforementioned normalization technique is necessary to address these issues.

\subsection{Alignment with KAN}
\begin{figure}[h!]
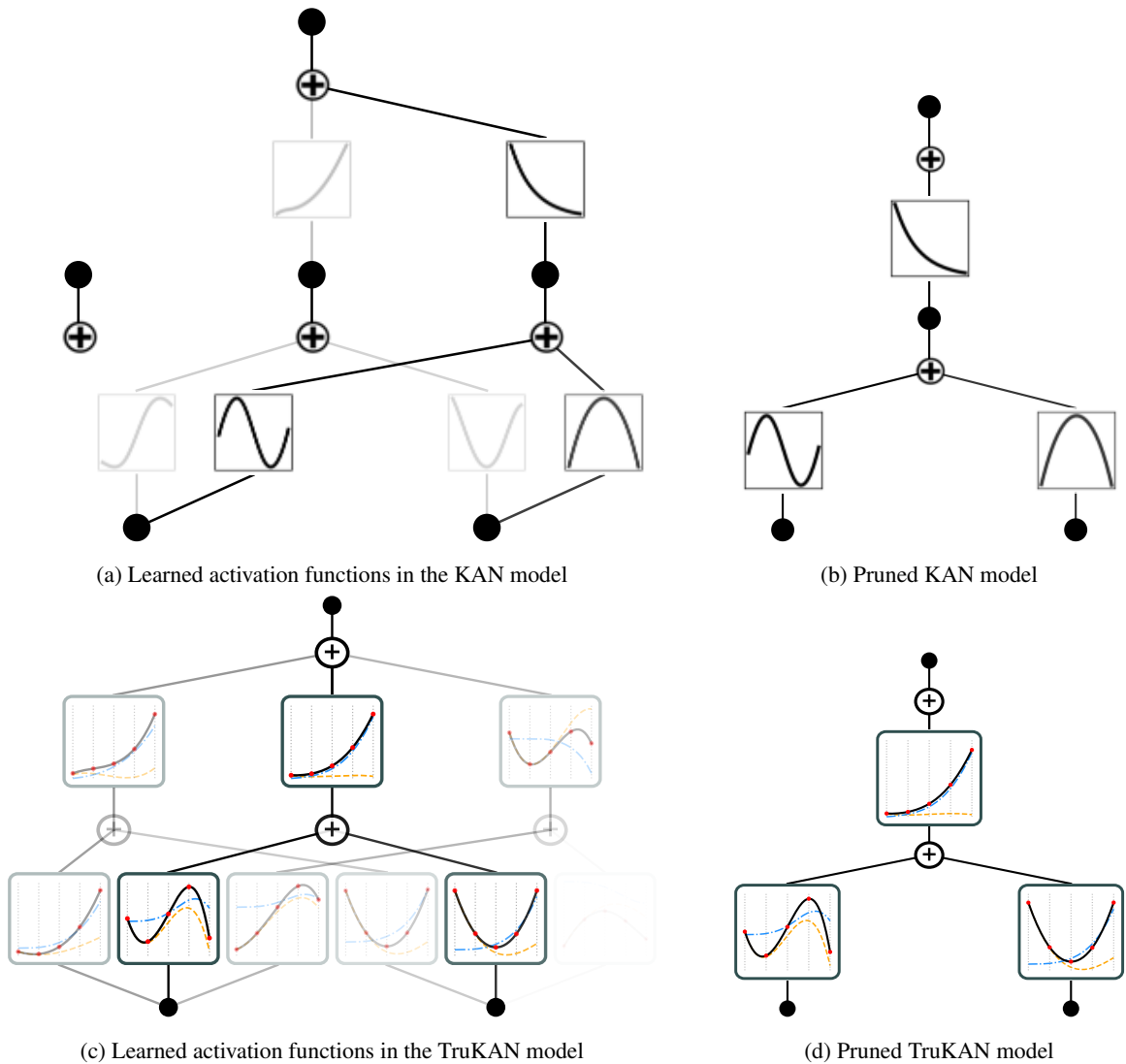

    \centering
    
    \begin{subfigure}[b]{0.60\textwidth}
        \centering
        \includesvg[inkscapelatex=false, width=0.92\linewidth]{images/align_kan_trained}
        \caption{Learned activation functions in the KAN model}
        \label{fig:align_kan_trained}
    \end{subfigure}
    \hfill
    \begin{subfigure}[b]{0.35\textwidth}
        \centering
        \includesvg[inkscapelatex=false, width=0.95\linewidth]{images/align_kan_pruned}
        \caption{Pruned KAN model}
        \label{fig:align_kan_pruned}
    \end{subfigure}
    \hfill
    \begin{subfigure}[b]{0.60\textwidth}
        \centering
        \includesvg[inkscapelatex=false, width=0.95\linewidth]{images/align_trukan_trained}
        \caption{Learned activation functions in the TruKAN model}
        \label{fig:align_trukan_trained}
    \end{subfigure}
    \hfill
    \begin{subfigure}[b]{0.35\textwidth}
        \centering
        \includesvg[inkscapelatex=false, width=0.98\linewidth]{images/align_trukan_pruned}
        \caption{Pruned TruKAN model}
        \label{fig:align_trukan_pruned}
    \end{subfigure}
    
    \caption{The illustration of KAN and TruKAN models, along with their pruned versions. (a) Learned activation functions in the KAN model; (b) Pruned KAN model after applying a magnitude threshold of 0.3; (c) Learned activation functions in the TruKAN model, where the composite activation (solid black) is decomposed into its polynomial term and truncated power term (dashed lines in distinct colors);  (d) Pruned TruKAN model after the same thresholding.}
    \label{fig:align_kan_trukan}
\end{figure}

KAN \cite{KAN,MultKAN} instantiates each univariate mapping as a mixture of a parametric activation (e.g. SiLU) and a spline correction term, and then sums those univariates across input coordinates to form each output channel. In KAN, the knot grid itself is initialized deterministically as a non-trainable parameter while the spline coefficients (the linear weights on the local spline basis) are learnable. Thus, this design decouples the placement of the basis functions in the input domain (fixed knots) from their combination (learned coefficients). From an engineering standpoint, this approach offers several advantages. First, evaluations can exploit pre-computed knot geometry, such as regular grids and repeated broadcast patterns. Second, the number of trainable parameters is restricted to coefficients and scales, reducing the risk of overfitting. Third, global nonlinearity (e.g., SiLU) reduces reliance on the spline to capture coarse trends. However, this design has some drawbacks. First, the fixed spline grid limits flexibility in representing knot locations to linear reweighting or grid expansion. Second, because the spline correction is usually implemented in a B-spline or other locally supported basis, attributing local changes to specific knot coefficients is more difficult than with bases that explicitly expose breakpoints contributions. In other words, while B-spline style bases possess compact, local support which tends to reduce the number of active basis functions at any input value and can improve numerical conditioning in large-knot regimes, their overlap complicates causal attribution of local shape changes to particular coefficients.

Compared to KAN, our TruKAN model improves flexibility and interpretability through its design choices. Overall, the TruKAN architecture utilized explicit decomposition of a low-order polynomial trend (the TruKAN basis component) and truncated-power functions that capture local deviations. This representation yields several practical design choices that alter the trade-offs.  First, truncation-based univariates make the contribution of each knot explicit and interpretable. In TruKAN, a non-zero coefficient at a knot $t_j$ indicates that the model intentionally introduces an additional k-th order curvature at precisely $t_j$, which greatly simplifies post-hoc inspection and scientific interpretation.   Second, TruKAN has learnable knot parameters, meaning that when the knot sequence $\{t_j\}_{j\in\mathbb{Z}}$ are permitted to move (subject to a numerically stable reparameterization that enforces ordering and monotonicity), the model can adapt the breakpoint locations to the data rather than being constrained by prior choices. It is worth noting that, while this added flexibility enhances representational power, it also requires careful regularization and robust parameterizations to avoid optimization problems and the resulting numerical instability during training.  

Third, TruKAN has explicit broadcasting control determined by one hyperparameter. This hyperparameter defines the broadcasting and whether a set of knots is shared across outputs (i.e., a shared position for knots) or if each output has its own set of knots. In case of fixed knots, the TruKAN structure will resemble the KAN models. For learnable knots, the position of the shared knots will be learned through all outputs. Shared knots not only optimize parameter count and encourage consistent structure across output dimensions, they also improve statistical strength and reduce the risk of overfitting in low-data regimes. Conversely, an individualized knot set enhances per-output expressivity at the cost of more learnable parameters and may diminish interpretability when numerous per-output basis sets are learned independently. 

Figure \ref{fig:align_kan_trained} and \ref{fig:align_trukan_trained} illustrate the KAN and TruKAN (with fixed knots) models trained for 1,000 iterations to approximate the target function $\mathrm{e}^{\sin(\pi x) + y^2}$. After training, both models are pruned using a threshold of 0.3 to obtain more compact representations (cf. \ref{fig:align_kan_pruned} and \ref{fig:align_trukan_pruned}). The results show that the two models achieve comparable approximations of the target function. The primary differences lie in the signs of the learned weights for the activation functions where the KAN model predominantly uses positive weights, whereas TruKAN employs negative weights. In the figures of TruKAN activations, the composite activation function is plotted in solid black to represent the primary curvature. The individual sub-components (the polynomial term and the truncated power term) are displayed in distinct colors using dashed lines for clear differentiation. Regarding network configuration and the number of learnable parameters, the KAN and TruKAN models are constructed using the same architecture, with a grid size of 3, a spline order of 3, and a layer configuration of three layers with 2, 3, and 1 neurons, respectively. The original KAN model has 108 learnable parameters, which reduces to 36 after pruning. The TruKAN has 67 learnable parameters before pruning, decreasing to 23 afterward. These results indicate that TruKAN achieves comparable representational capacity with substantially fewer parameters. All visualizations and pruning operations are performed using tools provided by the pykan \footnote{https://github.com/KindXiaoming/pykan} and TruKAN packages.

\section{Experiments Setup}
\subsection{Computer Vision Datasets}
This experiment part compares different network architectures (MLP-, KAN-, SineKAN- and TruKAN-based) using four benchmark image datasets, CIFAR-10, CIFAR-100, Oxford-IIIT Pets, and STL-10. CIFAR-10 and CIFAR-100 \cite{CIFAR} are standard benchmarks for visual recognition, containing low-resolution color images in 10 and 100 categories, respectively. Oxford-IIIT Pets \cite{OxfordPets} provides higher-resolution images of 37 breeds of pets, with significant intra-class variations due to different poses and lighting conditions. STL-10 \cite{STL10} is a medium-sized benchmark dataset drawn from ten semantically distinct classes to evaluate learning algorithms under realistic visual variability. The dataset provides a small labeled set of training and test images, and a large pool of unlabeled images. This evaluation is essential, as CIFAR-10 and CIFAR-100 serve as standardized benchmarks for fine-grained image classification under varying levels of class complexity. STL-10 focuses on supervised learning with limited labeled data, while Oxford-IIIT Pet assesses object detection performance amid natural variations in animal breeds and poses.
Table \ref{tab:vision_datasets} summarizes the main characteristics of these four datasets.  

\begin{table}[H]
\centering
\caption{Summary of datasets used in image classification task.}
\label{tab:vision_datasets}
\begin{tabular}{lcccc}
\toprule
\textbf{Dataset} & \textbf{Classes} & \textbf{Image Size} & \textbf{Training Samples} & \textbf{Test Samples} \\
\midrule
CIFAR-10         & 10  & $32 \times 32$ & 50,000  & 10,000 \\
CIFAR-100        & 100 & $32 \times 32$ & 50,000 & 10,000 \\
Oxford-IIIT Pets & 37  & Varying ($\sim$ $200 \times 200$) & 3,680 & 3,669 \\
STL-10           & 10  & $96 \times 96$ & 5,000 & 8,000 \\
\bottomrule
\end{tabular}
\end{table}

\subsection{Data Augmentation}
Moreover, we employ a comprehensive data augmentation pipeline to enhance model robustness and generalization on the image datasets by utilizing the Albumentations library \cite{Albumentations} for image transformations. The training transformations includes initial padding followed by random cropping, horizontal flipping, and small rotations (limited to ±15 degrees and applied with a probability of 0.3). To introduce variability in color and intensity, a one-of selection was used with a probability of $0.7$ between random brightness-contrast adjustments (limits of $0.2$) and hue-saturation-value shifts (limits of 10, 15, and 10, respectively). Additionally, subtle blurring or noise was incorporated via a one-of choice between Gaussian blur (limit $1-3$) and Gaussian noise (standard deviation range $0.14-0.343$, per-channel) with a probability of $0.25$. For regularization, a coarse dropout (similar to Cutout) was applied with a probability of $0.5$, creating a single rectangular hole covering $20-30\%$ of the image area filled with zeros. All images were normalized using dataset specific means and standard deviations. In contrast, the test transformations were minimal, consisting only of resizing and the same normalization. To further mitigate overfitting, we integrated advanced mixup techniques consist of CutMix \cite{Cutmix} and MixUp \cite{Mixup}, randomly selected one per batch. This augmentation strategy draws from established practices in vision benchmarks, promoting invariance to common perturbations while maintaining computational efficiency. To further mitigate overfitting, we integrate advanced MixUp techniques consisting of CutMix \cite{Cutmix} and MixUp \cite{Mixup}, with one technique randomly selected per batch.

\subsection{Model Optimization}
Prior KAN studies \cite{KAN, MultKAN} mostly used the L-BFGS optimizer for training. L-BFGS is a quasi-Newton method that provides rapid local convergence in a full-batch, low-dimensional regime. However, its reliance on accurate curvature estimates and line searches and its per-iteration memory and computational cost, makes it poorly suited to large models and stochastic mini-batch training. Therefore, we adopt a hybrid first-order optimization strategy as a practical and scalable alternative to L-BFGS. We optimize all the model parameters using the AdamW \cite{ADAMW} optimizer wrapped by the LookAhead meta-optimizer \cite{LookAhead}. AdamW provides adaptive learning rates for each parameter with decoupled weight decay, which stabilizes the training of deep networks under noisy gradient estimates and modern regularization schemes. LookAhead is a fast/slow weight scheme that periodically interpolates a slow anchor point with fast inner updates. Wrapping AdamW with LookAhead further smooths the optimization trajectory, reduces variance, and empirically improves convergence stability and generalization \cite{LookAhead}. This approach retains efficiency and mini-batch suitability necessary for scaling KAN architectures, while approximating the robust descent behavior provided by second-order methods. Additionally, we use parameter groups to provide fine-grained, role-aware optimization control over the learning parameters. 
\section{Model Development}
\subsection{EfficientNet-V2-based Framework}
\begin{figure}[h!]
\centering
        \includesvg[inkscapelatex=false, width=0.4\linewidth]{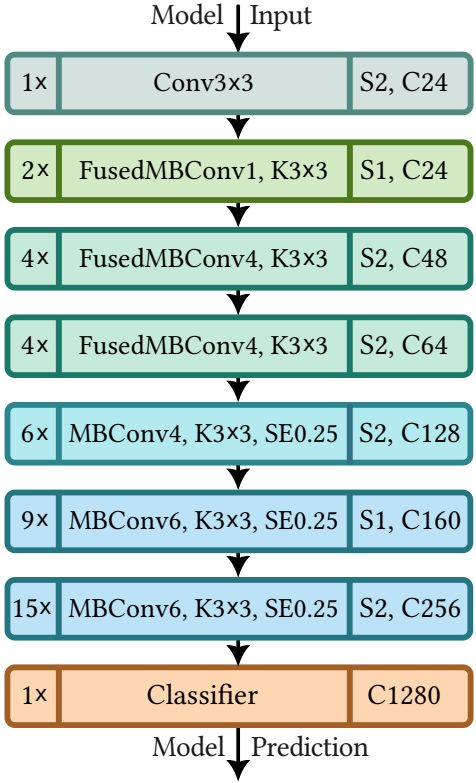}
        \caption{EfficientNet-V2 based framework to build eight different vision models.}
        \label{fig:EfficientNet_architecture}
\end{figure}

For vision tasks, we adopt an architecture based on the second variation of EfficientNet family \cite{Efficientnetv2}. The overall design is depicted in Figure \ref{fig:EfficientNet_architecture}. At the block level (the left side), the design incorporates two types of inverted bottleneck introduced in \cite{MobileNets}: (1) standard Mobile inverted convolution (MBConv) blocks, and (2) fused MBConv blocks. MBConv is a classic MobileNet-style inverted bottleneck consisting of pointwise and depthwise convolutions that expand and perform spatial filtering on the input channels. The SE factor (Squeeze-and-Excitation) defines the reduction ratio of the input channels. FusedMBConv is introduced as an alternative to MBConv for small and medium models, combining the depthwise (K×K) sequence, expansion and spatial filtering into a single regular convolution. The fused variant is used predominantly in the early layers, followed by the MBConv blocks. The left side of each block denotes the number of units, and the right side shows the number of channels and strides. 

For example, the block "$6\text{x} \:|\: \text{MBConv4 \;K}3\text{x}3, \text{SE}0.25 \:|\: \text{S2, C64}$", means a stack of six MBConvs units with a stride of 2 and a channel size of 64. The number 4 at the end of "MBConv4" refers to the expansion ratio of the inverted bottleneck convolution block, meaning the internal channel dimensions are expanded by a factor of 4 relative to the input channels during the expansion phase to increase model capacity while maintaining efficiency. "K3x3" indicates the kernel size of the convolution operation, specifically a 3x3 filter used in the depthwise or fused convolution layer, which balances receptive field and computational cost. Finally, "SE0.25" denotes the inclusion of a Squeeze-and-Excitation (SE) module with a reduction ratio of $0.25$. The SE mechanism adaptively recalibrates channel-wise feature responses by reducing the channel count to $25\%$ (or dividing by $4$) in its bottleneck layer before scaling back, providing an attention-like effect to enhance feature representation with minimal overhead. 

\begin{figure}[h!]
\centering
        \includesvg[inkscapelatex=false, width=0.7\linewidth]{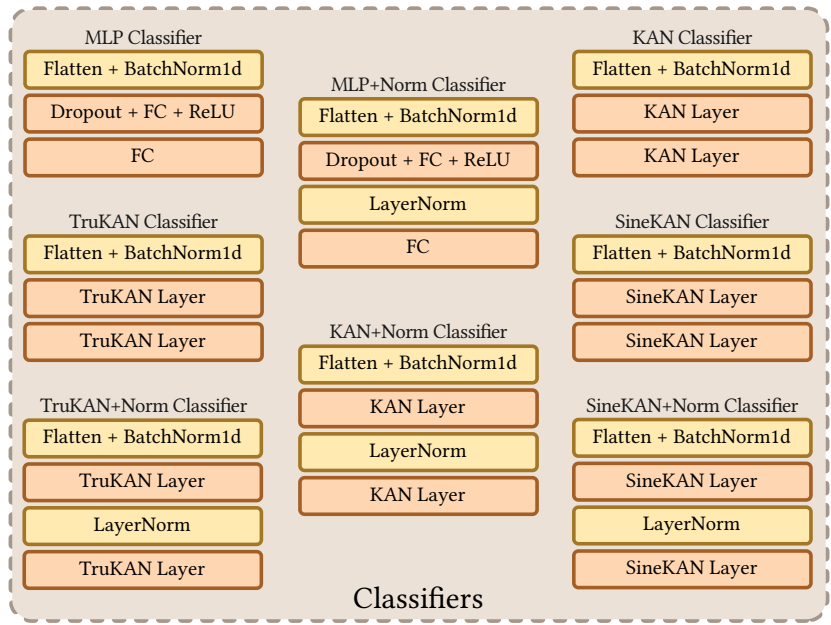}
        \caption{Selected classifiers: MLP, MLP+Normalization, KAN, KAN+Normalization,  SineKAN, SineKAn+Normalization, TruKAN and TruKAN+Normalization. }
        \label{fig:EfficientNet_classifier}
\end{figure}

The overall design of the feature map is consistent with the original EfficientNet-V2 design. The number of layers, channels, and strides are based on the configuration of a small EfficientNet variation. For comparison purposes, however, we replace the stock EfficientNet-V2 classifier (typically a single dropout followed by a linear layer that produces logits) 
with a new classifier consisting of a flatten layer (to flattens the input to yield a compact vector), a BatchNorm layer (to mitigates internal covariate shift by standardizing features across the mini-batch), and two basic layers (to generate the final logits) prefaced by the pooled features.

\subsection{Selected Classifiers}
As depicted in Figure \ref{fig:EfficientNet_classifier}, the two basic layers are based on one of the following components: 1) MLP Fully Connected (FC) layer (with a dropout layer followed by RELU layer), 2) TruKAN layer, 3) KAN layer and 4) SineKAN layer. We describe these classifiers as follows: 
\begin{itemize}
\item The MLP classifier consists of a standard two-layer fully connected network using ReLU as the activation functions, incorporating a dropout layer between the two fully connected layers. In the MLP variant with normalization, the layer normalization is applied after the dropout layer and before the second fully connected layer. The MLP, BatchNorm and LayerNorm components are implemented using PyTorch \cite{Pytorch}. 

\item The standard KAN implementation includes additional computational steps for visualization and pruning as part of the main training process. To ensure a fair comparison, we adopt the EfficientKAN implementation \footnote{\url{https://github.com/Blealtan/efficient-kan}}, which excludes these extra operations during training. We further optimize the model by vectorizing the B-spline basis function computations to improve computational efficiency. Considering the KAN's training process, the KAN implementation includes a dedicated mechanism to refine the grid based on the input data distribution. As this mechanism is unique to the KAN model, we exclude it from the model and training computations of the KAN classifier to ensure consistency with other models. 
\item For the SineKAN classifier, we adopt the official implementation provided by the authors. \footnote{\url{https://github.com/ereinha/SineKAN}}

\item The TruKAN classifier (which features a modular design with separate processes for pruning and visualization), we integrate the model directly into the training framework. In the normalized variants of all classifiers, a normalization layer is inserted between the two classifier layers.
\end{itemize}
It is noteworthy that differences in the internal architectures of the classifiers MLP, KAN, TruKAN and SineKAN naturally produce different parameter counts, so we adjust the number of neurons per layer for each model so that all models have approximately comparable numbers of trainable parameters. All other training hyperparameters are kept constant to ensure a fair comparison. In all experiments, we record training time and the number of trainable parameters, while the other performance metrics are evaluated on the test datasets. For each dataset, results are averaged over five training and testing trials. 
\subsection{Regularization}
Regarding generalization in KANs, prior studies have demonstrated efficacy and performance regression when applying conventional regularization techniques \cite{DropKAN,OurPpoKAN,ComSurveyKAN,OnTrainKAN}. The fundamental issue centers on how KANs differ architecturally from MLPs, requiring specialized generalization approaches rather than direct application of traditional regularization methods. In the KAN paper \cite{KAN, MultKAN}, entropy regularization was introduced to penalize duplicate activation functions and limit their number. This technique is effective at promoting sparsity among KAN neurons and spline coefficients, which makes networks simpler and more interpretable.
Considering weight decay regularization, empirical evidence \cite{ThreeWDecay,ComSurveyKAN} indicates that weight decay improves training dynamics during backpropagation. However, the effectiveness of weight decay depends on the choice of optimization algorithms, where particularly the weight decay technique provides strong performance boosts with second-order optimizers \cite{ThreeWDecay}.
Furthermore, data normalization can be used as an additional regularization mechanism. Techniques such as batch normalization \cite{BatchNorm} and layer normalization \cite{LayerNorm} stabilize training, accelerate convergence, and reduce internal covariate shift, and can provide regularization effects, reducing the need for more advanced regularization methods. 

Both normalization methods normalize activations to achieve zero mean and unit variance. However, they differ in the dimensions used for computing statistics.
Batch normalization computes the mean and variance across the mini-batch dimension for each feature independently, using data from the current batch. This dependency on batch statistics can cause training instability (especially with small batch sizes) and requires running averages for inference.
In contrast, layer normalization normalizes across the feature dimension for each training example. This approach is independent of the batch size parameter, which it appropriate for tasks with dynamic sequence lengths where batch statistics may be unreliable.

Furthermore, batch normalization and layer normalization are incorporated within each classifier modules. Additionally, we employ implicit L2 regularization via weight decay throughout the optimization process across all models.
\subsection{Hyperparameter Tuning}
Considering the distinct convergence behaviors of MLPs and KANs, we utilize a pre-training process to minimize the influence of different classifier architectures on the feature extraction process and ensure consistent comparison. Hence, an EfficientNet model is first trained with its original classification head. Next, alternative architectures are used for the classifier in comparative experiments, while the backbone weights are initialized from the pretrained model. In addition to the previously defined parameter groups, separate learning rates are applied to the backbone (feature extractor) and classifier modules. Model optimization follows the hybrid strategy (AdamW + LookAhead) with a weight decay of $10^{-4}$. The initial learning rate is set to $10^{-4}$ for backbone parameters and $10^{-3}$ for the classifier parameters.
To enhance training stability and convergence, we employ a learning rate scheduling strategy combined with a linear warm-up phase with cosine annealing decay. Specifically, during the first ten epochs, the learning rate is increased linearly from $\eta/10$ to $\eta$ using a warm-up scheduler. Following this stage, a cosine annealing scheduler gradually decreases the learning rate to a minimum of $10^{-5}$ across the remaining epochs, following the approach of \cite{SGDWarmRestarts}. Table \ref{tab:hyper_param} presents the hyperparameters of the experimental setup.

\begin{table}[h!]
  \centering
  \caption{Hyperparameters of the experiments}
      \begin{tabular}{l|l|l}
        \hline
        Name & Definition & Value \\ \hline
    
        Warmup epochs & \; Warmup steps for learning rate scheduler & 10 \\
        Batch size & \; Number of training samples processed simultaneously & 512 \\
        Initial learning rate & \; Initial value for step size during model training & 5e-4 \\
        Dropout & \; Dropout value for MLP models & 0.1 \\
        Order k & \; Order of splines & 3 \\
        Grid size & \; Number of points (knots) in grid mesh & 8\\
        Grid range & \; Range of point (knot) values & (-1.0, 1.0)\\ \hline
      \end{tabular}
  \label{tab:hyper_param}
\end{table}

\section{Model Evaluation}
In the following sections, we report in Tables \ref{tab:vision_results_s} and \ref{tab:vision_results_l} the training time, the number of trainable parameters, Accuracy and F1-score for each model across the four datasets. For all experiments, we try to keep the number of classifier parameters similar across methods to ensure that any performance differences is interpreted as algorithmic rather than due to capacity.
\subsection{Performance Evaluation on Small Networks} 

\begin{table}[h!]
\centering
\caption{Average Performance results on five trails on the small variant of the networks (in terms of number of parameters). Red numbers are the best results, and bold numbers are the runner-up results.}
\label{tab:vision_results_s}
    \begin{tabular}{l|ccc|ccc}
    \Xhline{3\arrayrulewidth}
    \multirow{2}{*}{Model} & \multicolumn{3}{c|}{CIFAR-10} & \multicolumn{3}{c}{STL-10}  \\
    \cline{2-7} & \makecell{Number of\\Parameter\\(Classifier)} & \makecell{F1 Score\\(Macro)} & Accuracy & \makecell{Number of\\Parameter\\(Classifier)} & \makecell{F1 Score\\(Macro)} & Accuracy \\ 
    \hline
    EfficientNet-MLP  & $\sim 333K$ & 0.8881 & 0.8888 & $\sim 333K$ & 0.7995 & 0.8025 \\
    EfficientNet-MLP-N  & $\sim 333K$ & 0.8901 & 0.8907 & $\sim 333K$ & {\color{red}0.9397} & {\color{red}0.9397} \\
    EfficientNet-KAN & $\sim 311K$ & 0.7714 & 0.7679 & $\sim 311K$ & 0.5743 & 0.5644 \\
    EfficientNet-KAN-N & $\sim 329K$ & 0.8781 & 0.8793 & $\sim 311K$ & 0.8391 & 8403 \\
    EfficientNet-SineKAN & $\sim 333K$ & 0.8775 & 0.8787 & $\sim 333K$ & 0.7767 & 0.7755 \\
    EfficientNet-SineKAN-N & $\sim 333K$ & 0.8802 & 0.8812 & $\sim 333K$ & 0.9225 & 0.9226 \\
    \Xhline{0.5\arrayrulewidth}
    EfficientNet-TruKAN-S & $\sim 304K$ & \textbf{0.8919} & \textbf{0.8925} & $\sim 304K$ & 0.7867 & 0.7848 \\
    EfficientNet-TruKAN-SN & $\sim 322K$ & 0.8908 & 0.8914 & $\sim 304K$ & \textbf{0.9340} & \textbf{0.9341} \\
    EfficientNet-TruKAN-I & $\sim 312K$ & 0.8672 & 0.8677 & $\sim 304K$ & 0.8333 & 0.8353 \\
    EfficientNet-TruKAN-IN & $\sim 312K$ & {\color{red}0.8922} & {\color{red}0.8927} & $\sim 304K$ & 0.9332 & 0.9333 \\
    \Xhline{2\arrayrulewidth}
    \multirow{2}{*}{Model} & \multicolumn{3}{c|}{CIFAR-100} & \multicolumn{3}{c}{Oxford-Pets}  \\
    \cline{2-7} & \makecell{Number of\\Parameter\\(Classifier)} & \makecell{F1 Score\\(Macro)} & Accuracy & \makecell{Number of\\Parameter\\(Classifier)} & \makecell{F1 Score\\(Macro)} & Accuracy \\ 
    \hline
    EfficientNet-MLP  & $\sim 350K$ & 0.6117 & 0.6198 & $\sim 340K$ & \textbf{0.7473} & \textbf{0.751} \\
    EfficientNet-MLP-N  & $\sim 356K$ & {\color{red}0.6306} & {\color{red}0.6370} & $\sim 340K$ & {\color{red}0.7477} & {\color{red}0.7514} \\
    EfficientNet-KAN & $\sim 331K$ & 0.0608 & 0.03 & $\sim 317K$ & 0.0818 & 0.0505 \\ 
    EfficientNet-KAN-N & $\sim 350K$ & 0.5171 & 0.5366 & $\sim 335K$ & 0.3112 & 0.3238 \\ 
    EfficientNet-SineKAN & $\sim 355K$ & 0.5296 & 0.545 & $\sim 340K$ & 0.635 & 0.6276 \\
    EfficientNet-SineKAN-N & $\sim 355K$ & 0.5546 & 0.5670 & $\sim 340K$ & 0.6396 & 0.6470 \\
    \Xhline{0.5\arrayrulewidth}
    EfficientNet-TruKAN-S & $\sim 316K$ & 0.5966 & 0.6070 & $\sim 308K$ & 0.681 & 0.6857 \\
    EfficientNet-TruKAN-SN & $\sim 344K$ & \textbf{0.6196} & \textbf{0.6274} & $\sim 329K$ & 0.7222 & 0.7261 \\
    EfficientNet-TruKAN-I & $\sim 334K$ & 0.4597 & 0.4843 & $\sim 319K$ & 0.4158 & 0.4301 \\
    EfficientNet-TruKAN-IN & $\sim 334K$ & 0.5711 & 0.5843 & $\sim 319K$ & 0.6950 & 0.6994 \\
    \Xhline{2\arrayrulewidth}
    \end{tabular}
\end{table}

Table \ref{tab:vision_results_s} summarizes the results of the small versions of the four classifiers. All results represent the average performance over five independent trials, with the primary metrics being the macro F1-score and Accuracy. The TruKAN variants incorporate either shared knots (with "-S" at their names) or individual knots per output (with "-I" at their names). As depicted in Figure \ref{fig:EfficientNet_classifier}, two variations are considered for each model, without or with layer normalization (marked with "N" at the end of their names).

Overall, layer normalization improved the performance of all the models. On the CIFAR-10 dataset, the TruKAN variants performed well where the TruKAN-IN variation achieved the highest macro F1-score and Accuracy, surpassing the best baseline models, underscoring the benefits of individual knots combined with layer normalization. The counterpart with shared knots, TruKAN-SN, yielded competitive results, nearly matching the top baseline. Without layer normalization, the TruKAN-SN model emerged as the runner-up, outperforming all other models except the normalized MLP variant while using slightly fewer parameters. In contrast, TruKAN-I lagged behind, suggesting a generalization problem that normalization effectively fixes in TruKAN-IN. Overall, the TruKAN variations with layer normalization consistently ranked at the top, indicating enhanced stability and generalization compared to the KAN and SineKAN baselines, which exhibited lower scores despite having similar parameter budgets.

For the STL-10 dataset, TruKAN variants again excelled, especially in normalized configurations. The TruKAN-SN model secured runner-up positions in both macro F1-score and Accuracy, trailing only the top baseline MLP-N by about $0.6\%$. Its individual-knots normalized counterpart, TruKAN-IN, performed similarly, demonstrating robustness across knot-sharing strategies when layer normalization is applied. Without normalization, TruKAN-S and TruKAN-I outperformed non-normalized baselines, like KAN and SineKAN, but fell short of normalized alternatives. Notably, all TruKAN models maintained a lower parameter count compared to baselines, highlighting their efficiency. These results affirm that TruKAN architectures, particularly with shared knots and normalization, provide substantial improvements over traditional KAN and SineKAN models, achieving near-state-of-the-art performance on this dataset.

\begin{table}[h!]
\centering
\caption{Average Performance Results on five trails on Large Networks Defined by the Number of Parameters. Red numbers are the best results, and bold numbers are the runner-up results.}
\label{tab:vision_results_l}
    \begin{tabular}{l|ccc|ccc}
    \Xhline{3\arrayrulewidth}
    \multirow{2}{*}{Model} & \multicolumn{3}{c|}{CIFAR-10} & \multicolumn{3}{c}{STL-10}  \\
    \cline{2-7} & \makecell{Number of\\Parameter\\(Classifier)} & \makecell{F1 Score\\(Macro)} & Accuracy & \makecell{Number of\\Parameter\\(Classifier)} & \makecell{F1 Score\\(Macro)} & Accuracy \\ 
    \hline
    EfficientNet-MLP  & $\sim 1.32M$ & 0.8890 & 0.8883 & $\sim 1.32M$ & 0.8057 & 0.8028 \\
    EfficientNet-MLP-N  & $\sim 1.32M$ & \textbf{0.8909} & \textbf{0.8913} & $\sim 1.32M$ & {\color{red}0.9376} & {\color{red}0.9377} \\
    EfficientNet-KAN & $\sim 2\:M$ & 0.8739 & 0.875 & $\sim 2\:M$ & 0.7332 & 0.7357 \\
    EfficientNet-KAN-N & $\sim 1.29\:M$ & 0.8791 & 0.8804 & $\sim 2\:M$ & 0.8785 & 0.8791 \\
    EfficientNet-SineKAN & $\sim 1.32M$ & 0.8685 & 0.8669 & $\sim 1.32M$ & 0.776 & 0.7770 \\
    EfficientNet-SineKAN-N & $\sim 1.32M$ & 0.8794 & 0.8804 & $\sim 1.32M$ & 0.9253 & 0.9254 \\
    \Xhline{0.5\arrayrulewidth}
    EfficientNet-TruKAN-S & $\sim 1.1M$ & 0.8695 & 0.8702 & $\sim 1.1M$ & 0.8262 & 0.8266 \\
    EfficientNet-TruKAN-SN & $\sim 1.25M$ & 0.8906 & 0.8901 & $\sim 1.1M$ & \textbf{0.9343} & \textbf{0.9344} \\
    EfficientNet-TruKAN-I & $\sim 1.1M$ & 0.8908 & \textbf{0.8913} & $\sim 1.1M$ & 0.7904 & 0.7916 \\
    EfficientNet-TruKAN-IN & $\sim 1.25M$ & {\color{red}0.8926} & {\color{red}0.8932} & $\sim 1.1M$ & 0.9332 & 0.9333 \\
    \Xhline{2\arrayrulewidth}
    \multirow{2}{*}{Model} & \multicolumn{3}{c|}{CIFAR-100} & \multicolumn{3}{c}{Oxford-Pets}  \\
    \cline{2-7} & \makecell{Number of\\Parameter\\(Classifier)} & \makecell{F1 Score\\(Macro)} & Accuracy & \makecell{Number of\\Parameter\\(Classifier)} & \makecell{F1 Score\\(Macro)} & Accuracy \\ 
    \hline
    EfficientNet-MLP  & $\sim 1.42M$ & 0.6037 & 0.6117 & $\sim 1.35M$ & \textbf{0.7477}  & \textbf{0.7516}  \\
    EfficientNet-MLP-N  & $\sim 1.42M$ & {\color{red}0.6321} & {\color{red}0.6389} & $\sim 1.35M$ & {\color{red}0.7482} & {\color{red}0.7517} \\
    EfficientNet-KAN & $\sim 2.14M$ & 0.5068 & 0.5278 & $\sim 2.04M$ & 0.4496 & 0.4639 \\ 
    EfficientNet-KAN-N & $\sim 1.37M$ & 0.5592 & 0.5709 & $\sim 1.31M$ & 0.6079 & 0.6145 \\
    EfficientNet-SineKAN & $\sim 1.42M$ & 0.5453 & 0.5561 & $\sim 1.35M$ & 0.6786 & 0.6821 \\
    EfficientNet-SineKAN-N & $\sim 1.42M$ & 0.5656 & 0.5765 & $\sim 1.35M$ & 0.6784 & 0.6825 \\
    \Xhline{0.5\arrayrulewidth}
    EfficientNet-TruKAN-S & $\sim 1.25M$ & 0.5225 & 0.5366 & $\sim 1.15M$ & 0.4726 & 0.4849 \\
    EfficientNet-TruKAN-SN & $\sim 1.34M$ & \textbf{0.6193} & \textbf{0.6256} & $\sim 1.28M$ & 0.7365 & 0.7397 \\
    EfficientNet-TruKAN-I & $\sim 1.25M$ & 0.6108 & 0.6189 & $\sim 1.15M$ & 0.71 & 0.712 \\
    EfficientNet-TruKAN-IN & $\sim 1.34M$ & 0.6123 & 0.6196 & $\sim 1.28M$ & 0.7369 & 0.7405 \\
    \Xhline{2\arrayrulewidth}
    \end{tabular}
\end{table}

About the challenging CIFAR-100 dataset, with finer-grained classes, TruKAN models showed significant advantages in handling complexity. The TruKAN variations with layer normalization performed well, where TruKAN-SN is the runner-up, nearly matching the top-performing MLP-N baseline and surpassing the other KAN variants. The normalized individual-knots model, TruKAN-IN, surpassed SineKAN-N and KAN-N. The non-normalized shared-knots model, TruKAN-S, exceeded all non-normalized baselines, including the drastically under-performing KAN model. However, TruKAN-I was less effective, which emphasizes the importance of good generalization with denser TruKAN architectures. Overall, the TruKAN variants delivered results that were superior to or comparable with the baselines (which had slightly higher parameters), illustrating their parameter efficiency and ability to mitigate issues in high-class scenarios.

On the Oxford-Pets dataset, the TruKAN models showed competitive performance, though the MLP baselines remained in the top spots. Among the TruKAN variants, TruKAN-SN led, outperforming SineKAN-N by over $8\%$ and KAN-N by a wide margin while approaching the runner-up MLP baseline. The normalized individual-knots TruKAN-IN and non-normalized shared-knots TruKAN-S significantly outperformed the KAN and SineKAN. However, TruKAN-I was an exception due to its non-normalized individual-knot design. Despite the top performance of the MLP baselines, the TruKAN models achieved these results with fewer parameters, suggesting potential for further optimization in domain-specific tasks.

Overall, across all datasets, the TruKAN architectures, especially those with shared knots and layer normalization (TruKAN-SN), consistently outperformed or matched leading baselines in efficiency and effectiveness. 
\subsection{Performance Evaluation on Large Networks} 
\label{large_networks}

Table \ref{tab:vision_results_l} reports the evaluation metrics for the larger variants of the four classifiers. On the CIFAR-10 dataset, 
the TruKAN variants take the top spots, with TruKAN-IN achieves the best performance by a clear margin, outperforming the MLP-N baseline while using fewer parameters. Remarkably, the TruKAN-SN and TruKAN-I variants also surpass or tie the best MLP baseline. Thus, three of the four TruKAN configurations are among the top four, demonstrating consistent superiority over the MLP, KAN and SineKAN at this scale.

The performance of the TruKAN variants remained consistent on the STL-10 dataset, where the TruKAN-SN variant delivered runner-up performance, trailing the leading MLP-N variant by only $0.3\%$, despite requiring $17\%$ fewer parameters. The normalized model, TruKAN-IN, followed closely behind. All of the TruKAN variants dramatically outperformed the KAN and SineKAN models, confirming that the TruKAN formulation scales more gracefully than other KAN models on this dataset.

On the challenging 100-class CIFAR benchmark, TruKAN-SN emerges once again as the runner-up, achieving results comparable to the best MLP-N baseline model. TruKAN-I and TruKAN-IN significantly outperform all KAN and SineKAN variants. These TruKAN variants exceed the strongest SineKAN-N result by over 5\%. The non-normalized TruKAN-S variant is competitive with earlier KAN-based models, but its normalized counterparts surpass it, reinforcing the benefit of layer normalization at larger scales.

On the Oxford-Pets dataset, TruKAN models demonstrate significant improvements in fine-grained recognition. The normalized variants, TruKAN-SN and TruKAN-IN, substantially outperform all previous KAN and SineKAN architectures. They approach the MLP baselines, with TruKAN-IN nearly matching the runner-up MLP performance while requiring fewer parameters.

Overall, scaling the classifier, which led to million parameters, amplifies the advantages of the proposed TruKAN architectures. Across all four datasets, at least one TruKAN variant achieves the best or second-best result. These variants consistently outperform vanilla KAN and SineKAN models by large margins, either surpassing or closely matching the strongest MLP baselines. However, the results of some TruKAN variants, such as TruKAN-S under the Oxford-Pets dataset, dropped detrimentally compared to their own results in smaller configurations with fewer parameters (cf. Table \ref{tab:vision_results_s}). These results suggest the generalization problem discussed previously. While MLP models can take advantage of more generalization techniques, such as dropout, KAN models perform poorly under some of these techniques.
\subsection{Training Time Comparison}
To evaluate the practical feasibility of our proposed TruKAN architectures, we conduct a comprehensive efficiency analysis, comparing them to baseline KAN variants across four metrics: total number of learnable parameters, average training time per step, GFLOPs (for the forward pass only), and peak GPU memory usage. Table \ref{tab:vision_time_s} shows the performance of the models for the CIFAR-100 dataset. The results are averaged (where applicable) over five trials. Red values denote the best performance in each metric and bold text indicates runner-ups. 

For this analysis, we employ two variants of the KAN model, differing in the treatment of the scale factors for the spline and SiLU basis functions. The KAN-PBT variant allows these scale factors to be learned during training. In contrast, the KAN-PBF variant uses fixed scale factors for both components, initialized to one divided by the square root of the input dimension. Similarly, for the TruKAN models, we consider two variants based on knot configurations. TruKAN-FS denotes the variant with fixed and shared knots across activations and TruKAN-FI employs fixed knots that are individual to each output activation.

The training step time reveals significant advantages for the shared-knot TruKAN models where TruKAN-LS secures second place at 0.0023 seconds, behind SineKAN's first place at 0.0018 seconds, with TruKAN-FS close behind at 0.0024 seconds. These times represent a 3-4x improvement over KAN-PBT and KAN-PBF models. In contrast, the TruKAN variants with individual knots for each output have higher latencies where TruKAN-FI has a latency of 0.0751 seconds, and TruKAN-LI has a latency of 0.1027 seconds. These latencies reflect the computational cost of output specific adaptations.

GFLOP \footnote{Giga Floating-Point Operations}, measures the model's computational complexity by calculating the required total number of floating-point arithmetic operations, typically for a single forward pass. Regarding the GFLOPs metric, the TruKAN models demonstrate a consistent performance of around 0.67 to 0.69. This is comparable to the KAN-PBF model (0.67) and is slightly above the best KAN-PBT model (0.53) and the runner-up, the SineKAN model (0.64). 

The memory usage further highlights TruKAN's strengths, with TruKAN-FS achieving the absolute best at 118.04 MB, outperforming SineKAN's runner-up 121.9 MB by approximately $3\%$. TruKAN-LS follows efficiently at 134.5 MB, offering a 4-5x reduction compared to KAN baselines (KAN-PBT at 570.84 MB and KAN-PBF at 691.79 MB). However, regarding TruKAN models with individual knot, these variants demand significantly more memory (TruKAN-FI at 2656.49 MB and TruKAN-LI at 2815.7 MB) due to the expanded knot storage per output.

Overall, across these metrics, the shared knot TruKAN configurations consistently emerge as the most efficient, often securing top or runner-up positions in time and memory while preserving competitive parameter and FLOPs profiles. 

\begin{table}[h!]
\centering
\caption{Efficiency comparison of KAN-based models across parameter count, average training step time, forward-pass GFLOPs, and peak GPU memory usage on the CIFAR-100 dataset. The values of time and memory are averaged across five trials. The red values indicate the best results, and bold denote runner-ups.}
\label{tab:vision_time_s}
    \begin{tabular}{lcccc}
    \toprule
    \textbf{Model} & \textbf{\makecell{Number of\\ Parameters}} & \textbf{\makecell{Avg Train \\Step Time (s)}} & \textbf{\makecell{GFLOPs\\(forward only)}} & \textbf{\makecell{Avg Peak GPU\\ Memory (MB)}} \\
    \midrule
        KAN-PBT & 5,227,200 & 0.0099 & {\color{red}0.53} & 570.84 \\
        KAN-PBF & 5,251,200 & 0.0087 & 0.67 & 691.79 \\
        SineKAN & 5,068,645 & {\color{red}0.0018} & \textbf{0.64} & \textbf{121.9} \\
        \Xhline{0.2\arrayrulewidth}
        TruKAN-FS & 5,279,440 & \textbf{0.0024} & 0.67 & {\color{red}118.04} \\
        TruKAN-FI & 5,279,440 & 0.0751 & 0.69 & 2656.49 \\   
    \bottomrule
    \end{tabular}
\end{table}

\section{Conclusions and Future Work}
We introduced a new architecture, TruKAN, to improve performance of KANs by preserving the canonical KAN topology and learnable activations. Our method replaces the recursive B-spline algorithm with a family of truncated power functions derived from k-order spline theory. We proposed two variants of TruKAN that differ in how the knots are used: shared knots over outputs or individual knots per output. We evaluated the performance of the TruKAN model on computer vision benchmark datasets using an advanced deep learning framework, and compared it with other models, such as MLP, KAN and SineKAN. Furthermore, to study the effect of the layer normalization within the classifiers, we deployed two variants of each model: one with normalization and one without. 
The results show the substitution B-spline by truncated power basis preserves the expressive power of KAN model, while delivering substantial improvements in Accuracy and F1-score over alternative variants. Also, the TruKAN model exhibits superior computational efficiency, consuming significantly less memory than other KAN variants and achieving comparable or reduced training times.

This research opens several avenues for future investigation. Regarding model generalization, the present study used weight decay and normalization techniques for TruKAN models. Subsequent work will examine strategies for integrating and stabilizing additional methods, such as dropout, to mitigate overfitting which preliminary evidence of it has been observed. In our work, all the results were obtained using 32-bit floating-point precision. Nevertheless, recent studies indicate that mixed-precision computations can enhance training efficiency without compromising the accuracy of the model. Therefore, future efforts will explore stabilizing KAN models under mixed-precision operations. Considering the KAN's interpretable edge-univariate functions, we will also demonstrate how TruKAN can effectively advance the field of Explainable AI through concrete applications. We are also interested in adopting TruKAN in other complex environments where MPLs may struggle, such as reinforcement learning and time-series forecasting. 

\bibliographystyle{alpha}
\bibliography{references}

\newcommand{\etalchar}[1]{$^{#1}$}
\begin{thebibliography}{QMW{\etalchar{+}}25}

\bibitem[Alt24]{DropKAN}
Mohammed~Ghaith Altarabichi.
\newblock {DropKAN: Regularizing kans by masking post-activations}.
\newblock {\em arXiv preprint arXiv:2407.13044}, 2024.

\bibitem[BC24]{WaveKAN}
Zavareh Bozorgasl and Hao Chen.
\newblock {WAV-KAN: Wavelet Kolmogorov-arnold Networks}.
\newblock {\em arXiv preprint arXiv:2405.12832}, 2024.

\bibitem[BIK{\etalchar{+}}20]{Albumentations}
Alexander Buslaev, Vladimir~I. Iglovikov, Eugene Khvedchenya, Alex Parinov, Mikhail Druzhinin, and Alexandr~A. Kalinin.
\newblock {Albumentations: Fast and Flexible Image Augmentations}.
\newblock {\em Information}, 11(2), 2020.

\bibitem[BKH16]{LayerNorm}
Jimmy~Lei Ba, Jamie~Ryan Kiros, and Geoffrey~E. Hinton.
\newblock {Layer Normalization}.
\newblock In {\em arXiv preprint arXiv:1607.06450}, 2016.

\bibitem[BMS25]{OurSacKAN}
Ali Bayeh, Malek Mouhoub, and Samira Sadaoui.
\newblock {Enhancing Off-Policy Method SAC with KAN for Continuous Reinforcement Learning}.
\newblock In {\em {Proc. of International Conference on Deep Learning Theory and Applications, DeLTA}}, pages 229--239, Cham, 2025. Springer.

\bibitem[BMS26]{OurPpoKAN}
Ali Bayeh, Malek Mouhoub, and Samira Sadaoui.
\newblock {KAN-based On-policy PPO Method for Continuous Reinforcement Learning}.
\newblock In {\em {Dorronsoro, B., Talbi, EG., Weraikat, D., Bouamama, S. (eds) Optimization and Learning. OLA 2025. Communications in Computer and Information Science, vol 2808.}}, 2026.

\bibitem[Che24]{visionKAN}
Minjong Cheon.
\newblock {Demonstrating the efficacy of kolmogorov-arnold networks in vision tasks}.
\newblock In {\em arXiv preprint arXiv:2406.14916}, 2024.

\bibitem[CNL11]{STL10}
Adam Coates, Andrew~Y. Ng, and Honglak Lee.
\newblock {An Analysis of Single-Layer Networks in Unsupervised Feature Learning}.
\newblock In {\em {Proceedings of the Fourteenth International Conference on Artificial Intelligence and Statistics (AISTATS)}}, pages 215--223, 2011.

\bibitem[CP{\etalchar{+}}24]{GenomicKAN}
Oleksandr Cherednichenko, Maria Poptsova, et~al.
\newblock {Kolmogorov-Arnold Networks for Genomic Tasks}.
\newblock {\em bioRxiv}, 2024.
\newblock bioRxiv preprint / Briefings in Bioinformatics (see repository).

\bibitem[dB78]{deboor1978}
Carl de~Boor.
\newblock {\em {A Practical Guide to Splines}}, volume~27 of {\em {Applied Mathematical Sciences}}.
\newblock Springer-Verlag, New York, 1978.

\bibitem[HDW{\etalchar{+}}20]{LightGCN}
Xiangnan He, Kuan Deng, Xiang Wang, Yan Li, Yongdong Zhang, and Meng Wang.
\newblock {LightGCN: Simplifying and Powering Graph Convolution Network for Recommendation}.
\newblock In {\em Proceedings of the 43rd International ACM SIGIR conference on research and development in Information Retrieval}, pages 639--648, 2020.

\bibitem[HZC{\etalchar{+}}17]{MobileNets}
Andrew~G. Howard, Menglong Zhu, Bo~Chen, Dmitry Kalenichenko, Weijun Wang, Tobias Weyand, Marco Andreetto, and Hartwig Adam.
\newblock {MobileNets: Efficient Convolutional Neural Networks for Mobile Vision Applications}.
\newblock In {\em arXiv:1704.04861}, 2017.

\bibitem[IS15]{BatchNorm}
Sergey Ioffe and Christian Szegedy.
\newblock {Batch normalization: accelerating deep network training by reducing internal covariate shift}.
\newblock In {\em Proceedings of the 32nd International Conference on International Conference on Machine Learning - Volume 37}, ICML'15, page 448–456. JMLR.org, 2015.

\bibitem[JHZ24]{ComSurveyKAN}
Tianrui Ji, Yuntian Hou, and Di~Zhang.
\newblock {A Comprehensive Survey on Kolmogorov Arnold Networks (KAN)}.
\newblock {\em arXiv:2407.11075v7 [cs.LG]}, 2024.

\bibitem[KGW{\etalchar{+}}24]{biomedicalKAN}
William Knottenbelt, Zeyu Gao, Rebecca Wray, Woody~Zhidong Zhang, Jiashuai Liu, and Mireia Crispin-Ortuzar.
\newblock {CoxKAN: Kolmogorov--Arnold Networks for Interpretable, High-Performance Survival Analysis}.
\newblock {\em arXiv:2409.04290}, 2024.

\bibitem[KKD25]{LeanKAN}
Benjamin~C. Koenig, Suyong Kim, and Sili Deng.
\newblock {LeanKAN: a parameter-lean Kolmogorov-Arnold network layer with improved memory efficiency and convergence behavior}.
\newblock {\em Neural Networks}, 192:107883, 2025.

\bibitem[Kri09]{CIFAR}
Alex Krizhevsky.
\newblock {Learning Multiple Layers of Features from Tiny Images}.
\newblock Technical report, University of Toronto, 2009.
\newblock Technical Report.

\bibitem[LH17]{SGDWarmRestarts}
Ilya Loshchilov and Frank Hutter.
\newblock {SGDR: Stochastic Gradient Descent with Warm Restarts}.
\newblock In {\em International Conference on Learning Representations (ICLR)}, 2017.

\bibitem[LH19]{ADAMW}
Ilya Loshchilov and Frank Hutter.
\newblock {Decoupled Weight Decay Regularization}.
\newblock In {\em International Conference on Learning Representations (ICLR)}, 2019.

\bibitem[LMW{\etalchar{+}}24]{MultKAN}
Ziming Liu, Pingchuan Ma, Yixuan Wang, Wojciech Matusik, and Max Tegmark.
\newblock {KAN 2.0: Kolmogorov-arnold networks meet science}.
\newblock {\em arXiv preprint arXiv:2408.10205}, 2024.

\bibitem[LWV{\etalchar{+}}25]{KAN}
Ziming Liu, Yixuan Wang, Sachin Vaidya, Fabian R{\"u}hle, James Halverson, Marin Solja\v{c}i\'c, Thomas~Y. Hou, and Max Tegmark.
\newblock {KAN: Kolmogorov-Arnold Networks}.
\newblock In {\em Proceedings of the International Conference on Learning Representations (ICLR)}, 2025.
\newblock ICLR 2025.

\bibitem[LZQL24]{FinanceKAN}
Charles~Z. Liu, Ying Zhang, Lu~Qin, and Yongfei Liu.
\newblock {Kolmogorov-Arnold Finance-Informed Neural Network in Option Pricing/KAFIN}.
\newblock {\em {Applied Sciences}}, 14(24):11618, 2024.

\bibitem[OPI{\etalchar{+}}25]{FinanceKAN2}
Vidhi Oad, Param Pathak, Nouhaila Innan, Muhammad Shafique, et~al.
\newblock {KASPER: Kolmogorov Arnold Networks for Stock Prediction and Explainable Regimes}.
\newblock {\em arXiv preprint arXiv:2507.18983}, 2025.

\bibitem[PGM{\etalchar{+}}19]{Pytorch}
Adam Paszke, Sam Gross, Francisco Massa, Adam Lerer, James Bradbury, Gregory Chanan, Trevor Killeen, Zeming Lin, Natalia Gimelshein, Luca Antiga, et~al.
\newblock {Pytorch: An imperative style, high-performance deep learning library}.
\newblock {\em Advances in neural information processing systems}, 32, 2019.

\bibitem[PVZJ12]{OxfordPets}
Omkar~M Parkhi, Andrea Vedaldi, Andrew Zisserman, and C.~V. Jawahar.
\newblock {Cats and Dogs}.
\newblock {\em 2012 IEEE Conference on Computer Vision and Pattern Recognition}, pages 3498--3505, 2012.

\bibitem[QMW{\etalchar{+}}25]{PowerMLP}
Ruichen Qiu, Yibo Miao, Shiwen Wang, Yifan Zhu, Lijia Yu, and Xiao-Shan Gao.
\newblock {PowerMLP: An Efficient Version of KAN}.
\newblock {\em Proceedings of the AAAI Conference on Artificial Intelligence}, 39(19):20069–20076, April 2025.

\bibitem[RRG25]{SineKAN}
Eric Reinhardt, Dinesh Ramakrishnan, and Sergei Gleyzer.
\newblock {SineKAN: Kolmogorov-Arnold Networks using sinusoidal activation functions}.
\newblock {\em Frontiers in Artificial Intelligence}, 7:1462952, 2025.

\bibitem[Sch07]{Schumaker2007}
Larry Schumaker.
\newblock {\em {Spline Functions: Basic Theory}}.
\newblock Cambridge University Press, August 2007.

\bibitem[Soh24]{OnTrainKAN}
Shairoz Sohail.
\newblock {On Training of Kolmogorov-Arnold Networks}.
\newblock {\em arXiv.org arXiv:2411.05296}, 2024.

\bibitem[Ta25]{BSRBFKAN}
Hoang-Thang Ta.
\newblock {BSRBF-KAN: A Combination of B-Splines and Radial Basis Functions in Kolmogorov-Arnold Networks}.
\newblock In Wray Buntine, Morten Fjeld, Truyen Tran, Minh-Triet Tran, Binh Huynh Thi~Thanh, and Takumi Miyoshi, editors, {\em Information and Communication Technology}, pages 3--15, Singapore, 2025. Springer Nature Singapore.

\bibitem[TL21]{Efficientnetv2}
Mingxing Tan and Quoc Le.
\newblock {Efficientnetv2: Smaller models and faster training}.
\newblock In {\em International conference on machine learning}, pages 10096--10106, 2021.

\bibitem[XCL{\etalchar{+}}24]{FourierKAN}
Jinfeng Xu, Zheyu Chen, Jinze Li, Shuo Yang, Wei Wang, Xiping Hu, and Edith C-H Ngai.
\newblock {FourierKAN-GCF: Fourier Kolmogorov-Arnold Network--An Effective and Efficient Feature Transformation for Graph Collaborative Filtering}.
\newblock {\em arXiv preprint arXiv:2406.01034}, 2024.

\bibitem[YHO{\etalchar{+}}19]{Cutmix}
Sangdoo Yun, Dongyoon Han, Seong~Joon Oh, Sanghyuk Chun, Junsuk Choe, and Youngjoon Yoo.
\newblock {Cutmix: Regularization strategy to train strong classifiers with localizable features}.
\newblock In {\em Proceedings of the IEEE/CVF international conference on computer vision}, pages 6023--6032, 2019.

\bibitem[ZCDLP17]{Mixup}
Hongyi Zhang, Moustapha Cisse, Yann~N Dauphin, and David Lopez-Paz.
\newblock {Mixup: Beyond Empirical Risk Minimization}.
\newblock {\em arXiv preprint arXiv:1710.09412}, 2017.

\bibitem[ZLBH19]{LookAhead}
Michael Zhang, James Lucas, Jimmy Ba, and Geoffrey~E Hinton.
\newblock {Lookahead Optimizer: k steps forward, 1 step back}.
\newblock {\em Advances in neural information processing systems}, 32, 2019.

\bibitem[ZSMX25]{MetaKAN}
Zhangchi Zhao, Jun Shu, Deyu Meng, and Zongben Xu.
\newblock {Improving Memory Efficiency for Training KANs via Meta Learning}.
\newblock {\em arXiv preprint arXiv:2506.07549}, 2025.

\bibitem[ZWXG18]{ThreeWDecay}
Guodong Zhang, Chaoqi Wang, Bowen Xu, and Roger~Baker Grosse.
\newblock {Three Mechanisms of Weight Decay Regularization}.
\newblock {\em ArXiv:1810.12281}, abs/1810.12281, 2018.

\end{thebibliography}
\end{document}